\pdfoutput=1
\documentclass[runningheads]{llncs}
\usepackage[T1]{fontenc}
\usepackage{amsmath}
\usepackage{newtxtext}       %
\usepackage[varvw]{newtxmath}       

\usepackage{graphicx}
\usepackage{booktabs}
\usepackage{makecell}        
\makeatletter
\g@addto@macro\normalsize{%
  \setlength{\abovedisplayskip}{7pt plus 2pt minus 3pt}%
  \setlength{\belowdisplayskip}{7pt plus 2pt minus 3pt}%
  \setlength{\abovedisplayshortskip}{2pt plus 1pt}%
  \setlength{\belowdisplayshortskip}{4pt plus 1pt minus 1pt}}
\makeatother
\newcommand{\pmsd}[2]{$#1\,{\scriptstyle\pm\,#2}$}
\usepackage[hidelinks]{hyperref}
\usepackage{color}

\begin{document}
\title{BAM: Bivariate Additive Models for Interpretable Tabular Prediction}
\titlerunning{BAM: Bivariate Additive Models}
%
%
\author{Atsushi Koike\orcidID{0000-0002-6262-4308}}
\authorrunning{A. Koike}
\institute{Tohoku University, Sendai, Japan\\
\email{koike@tohoku.ac.jp}}
\maketitle              
\begin{abstract}
Generalized Additive Models with pairwise interactions (GA\textsuperscript{2}Ms) are regarded as glass-box models, since each pairwise shape function can be visualized as a contour plot. In practice, the shape functions learned by methods such as EBM or NODE-GAM are so complex---irregular contours and sharp local fluctuations---that a practitioner cannot meaningfully interpret them or check them against domain knowledge. Auditability in this stronger sense requires substantially simpler shape functions. We argue that such simplification can match the accuracy of standard GAMs if the right feature pairs are captured. We propose Bivariate Additive Models (BAMs), a class of GA\textsuperscript{2}Ms in which every pairwise shape function is constrained to a simple parametric form, instantiated as an asymmetric bivariate Gaussian. Each component is visualizable as a unimodal contour map, and its feature pair is selected automatically during training. Experiments on six tabular benchmarks show that BAM attains predictive accuracy comparable to standard GAMs and approaches that of GA\textsuperscript{2}Ms. We also show the trade-off between accuracy and the number of components to inspect.

\end{abstract}
\section{Introduction}
\label{sec:intro}

Machine learning is widely used for important decisions in society, such as medical diagnosis \cite{caruana15}, financial lending, and judicial judgment \cite{rudin19}. A closely related need arises in industrial settings (e.g., a production line monitored by many sensors), where practitioners must identify which small subset of variables---or pairs of variables---drives an outcome from among hundreds of recorded signals, in a form they can quickly share with domain experts. In such settings, high predictive performance alone is insufficient: the criteria behind specific predictions must be transparent and intelligible enough to be examined, questioned, and accepted as reasonable.

Unfortunately, understanding why a particular prediction was made is typically challenging. Techniques such as LIME \cite{ribeiro16} and SHAP \cite{lundberg17} have been proposed to provide post-hoc explanations for predictions. However, these explanations can differ substantially from actual model behavior \cite{slack20}. Models whose prediction processes are inherently visible are known as glass-box models. Against this backdrop, the importance of glass-box models is increasingly recognized.

Generalized Additive Models (GAMs) are a representative class of glass-box models that achieve high predictive accuracy.
GAMs compute a score for each feature and sum them up along with a bias term to obtain the final prediction.
For example, when predicting disease risk from health checkup data, the model uses a per-feature shape function to compute an individual risk score for each measurement (e.g., blood pressure, BMI), and these individual scores are summed to obtain the total risk score.
Visualizing these shape functions provides interpretability of the prediction process.
GA\textsuperscript{2}M extends this concept by incorporating pairwise interactions. In GA\textsuperscript{2}Ms, shape functions can depend on two features (e.g., a function of both ``age'' and ``blood pressure''), which allows the model to capture complex relationships between features.

Existing research on GAMs and GA\textsuperscript{2}Ms has primarily focused on approximating the true underlying function as accurately as possible.
Consequently, the resulting shape functions tend to become excessively complex (as shown in Fig.~\ref{fig:ebm_visualization}), exhibiting irregular contours and sharp local fluctuations.
Such complexity makes it difficult for a practitioner to interpret each shape function on its own. Checking whether the criterion the model has learned makes sense given established knowledge in the application domain is even more difficult. For applications where the model itself must be auditable---not merely accurate---these complex shape functions are unsuitable, regardless of how well they approximate the underlying function.

To address this, we argue that the shape functions of a GA\textsuperscript{2}M should be substantially simpler than those produced by unconstrained fitting. A natural concern is that such radical simplification will sacrifice predictive accuracy. The central claim of this paper is that this is not necessarily true: if the right pairs of features are captured, strongly constrained pairwise functions can match the accuracy of standard GAMs and approach that of GA\textsuperscript{2}Ms.

We propose Bivariate Additive Models (BAMs), a class of GA\textsuperscript{2}Ms in which every pairwise shape function is constrained to a simple parametric form. BAM learns the feature pair for each component via a differentiable selection mechanism. Each component is visualizable as a single contour map. In this work, we instantiate the shape function as an asymmetric bivariate Gaussian (Bigaussian).

\begin{figure}[t]
    \centering
    \begin{minipage}[b]{0.29\linewidth}\centering
        \includegraphics[width=\linewidth]{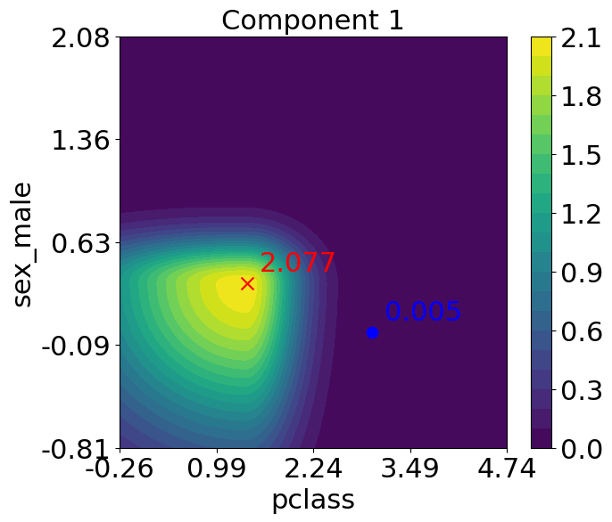}
    \end{minipage}\hfill
    \begin{minipage}[b]{0.29\linewidth}\centering
        \includegraphics[width=\linewidth]{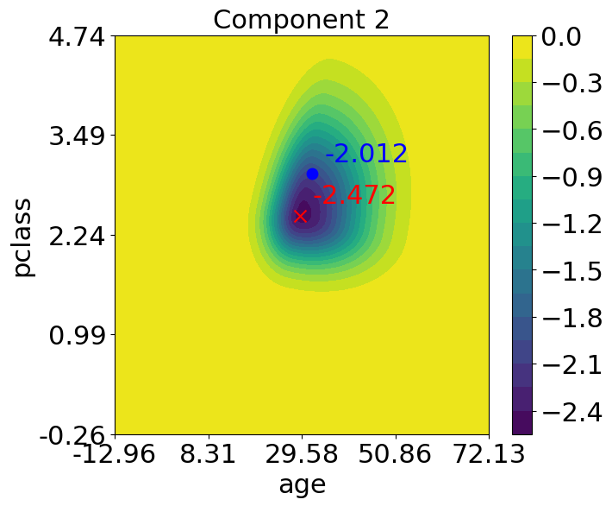}
    \end{minipage}\hfill
    \begin{minipage}[b]{0.29\linewidth}\centering
        \includegraphics[width=\linewidth]{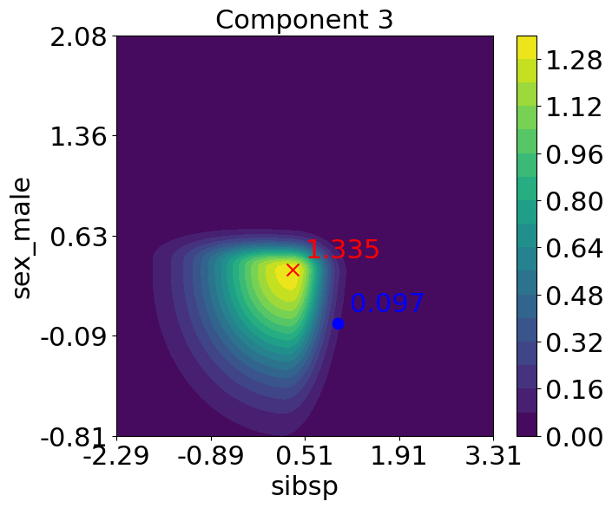}
    \end{minipage}

    \begin{minipage}{\linewidth}
    \small
    \begin{align*}
    \text{logit} &= \text{bias} + \text{comp.1} + \text{comp.2} + \text{comp.3} = (-0.010) + (0.005) + (-2.012) + (0.097) = -1.919 \\
    &< 0 \Rightarrow \text{prediction} = 0 \quad (\text{true label} = 0)
    \end{align*}
    \end{minipage}

    \caption{Visualization of passenger survival prediction on the Titanic dataset using the proposed model (BAM). The model calculates a single prediction by summing the scores from three bivariate graphs and a bias term ($-0.010$). The visualized values represent logits, where a positive total score indicates a prediction of survival. The blue points indicate the scores for a single sample, and the red crosses represent the centers of the Gaussian components. 
    }
    \label{fig:visualization}
\end{figure}

BAM computes the prediction as a sum of bivariate functions.
This simplicity allows users to intuitively understand how input values affect the output.
Figure~\ref{fig:visualization} shows an example of BAM applied to the Titanic passenger survival prediction task.
In this example, the model calculates a logit by summing the outputs of three components (bivariate functions) and a bias term ($-0.010$). The survival probability is obtained by passing this logit through a sigmoid function, where a positive logit indicates a prediction of survival.
Crucially, important features for prediction are aggregated into these three components.
For instance, the first component indicates that the survival probability increases for females (\texttt{sex\_male}=0) in the upper class (\texttt{pclass}=1).
In this way, features important for prediction are successfully extracted into a few components.

Each component in BAM represents the deviation of the output from a baseline value (bias term).
The concept of attributing deviations from a baseline to features is similar to SHAP \cite{lundberg17}.
However, while SHAP estimates feature contributions in a post-hoc manner, BAM directly adds the calculated values to the final output.
Consequently, BAM is inherently free from the reliability issues associated with post-hoc contribution estimation in SHAP.
Furthermore, BAM can assign values to feature pairs, which is particularly effective for datasets where interactions are critical.

We empirically demonstrate that BAMs achieve predictive accuracy comparable to standard GAMs and approach that of GA\textsuperscript{2}Ms. We also compare BAM visualizations with those of GA\textsuperscript{2}Ms and SHAP, and show that BAM makes the overall model behavior easier to grasp. Furthermore, we show how accuracy improves as the number of components increases.

\section{Preliminaries}
\label{sec:preliminaries}

\subsection{GAM}
\label{sec:GAM}

Among glass-box models, the Generalized Additive Model (GAM) is widely known for achieving high predictive performance while maintaining interpretability. GAM calculates a score for each input feature and obtains the final output by summing these scores together with a bias term.

Given $m$ features $\boldsymbol{x} = [x_1, x_2, \dots, x_m]$, the GAM prediction for the target value, denoted by $y$, is represented as:
\begin{equation}
g(y) = \beta_0 + \sum_{i=1}^{m} f_i(x_i),
\end{equation}
where $g$ is a link function (e.g., logit for classification), and $\beta_0$ represents a bias term.
For instance, in disease risk prediction, each feature $x_i$ (e.g., BMI) yields a risk score $f_i(x_i)$, and the total risk is determined by summing these scores and the bias term $\beta_0$.

A key challenge in GAMs is the estimation of the shape functions $f_i$. Explainable Boosting Machine (EBM) \cite{caruana15,lou12} utilizes boosted decision trees to estimate these functions, thereby achieving predictive performance comparable to black-box models. More recently, deep learning-based approaches have been proposed to learn each shape function more flexibly: Neural Additive Models (NAM) \cite{rishabh21} use one sub-network per feature; NODE-GAM \cite{chang22} adopts oblivious decision trees with differentiable feature selection; Neural Basis Models (NBM) \cite{radenovic2022} share basis functions across features for scalability; LA-NAM \cite{bouchiat2024} introduces Bayesian principles for uncertainty estimation; and GP-NAM \cite{zhang2024} uses a single-layer neural-network construction of Gaussian processes via random Fourier features, yielding a convex objective. These methods improve expressivity in various ways; however, the learned shape functions remain free-form rather than constrained to a compact parametric form, which can make them difficult to interpret on their own. BAM differs in design philosophy by constraining each (pairwise) shape function to a compact parametric form rather than maximizing flexibility.

\subsection{GA\textsuperscript{2}M}
\label{sec:GA2M}

GA\textsuperscript{2}M (Generalized Additive Models with pairwise interactions)~\cite{lou13} is an extension of GAMs that incorporates pairwise feature interactions. The GA\textsuperscript{2}M prediction is formulated as:
\begin{equation}
g(y) = \beta_0 + \sum_{i=1}^{m} f_i(x_i) + \sum_{i=1}^{m-1}\sum_{j=i+1}^{m} f_{i,j}(x_i,x_j).
\end{equation}
By explicitly modeling pairwise feature interactions through $f_{i,j}$, GA\textsuperscript{2}Ms can achieve higher predictive performance than standard GAMs. However, incorporating these pairwise functions, each of which has a potentially complex shape, makes the resulting model hard to interpret intuitively. Note that it is not necessary to include interactions for every feature pair; unimportant interactions can be omitted by setting $f_{i,j}=0$.

Modern implementations, such as EBM and NODE-GAM, support this GA\textsuperscript{2}M framework to enhance predictive performance. More recently, GRAND-SLAMIN' \cite{ibrahim2023} addresses the combinatorial growth of pairwise terms in GA\textsuperscript{2}M by imposing structural constraints on which feature pairs are included, learning sparse selections of pairs through differentiable optimization on tree-based shape functions. Like GRAND-SLAMIN', BAM learns feature pairs end-to-end, 
but additionally constrains the form of each pairwise function to a parametric ellipse, which yields visualizations whose geometric semantics can be read directly from learned parameters.

\section{Proposed Model}
\label{sec:our_method}

\subsection{Bivariate Additive Model}
BAMs are a class of GA\textsuperscript{2}Ms. They constrain every pairwise shape function to a simple parametric form.
We first introduce the general form of BAMs and then describe the specific shape functions used in this study.

Let $\boldsymbol{x} = [x_1, x_2, \dots, x_m] \in \mathbb{R}^m$ be the input features and $y$ be the target. We define a tuple of feature index pairs $\mathcal{I}$ to specify the input features for $n_c$ bivariate shape functions:
\begin{equation} \label{eq:feature_indecies}
    \mathcal{I} = \left[~(i[1], j[1]), \dots, (i[n_c], j[n_c])~\right].
\end{equation}
The BAM prediction is defined as:
\begin{equation} \label{eq:bam_score}
    g(y) = \beta_0 + \sum_{k=1}^{n_c} f_k(x_{i[k]},x_{j[k]}),
\end{equation}
where $g(\cdot)$ is a link function and $f_k$ is the $k$-th bivariate shape function. For regression tasks, we use the identity link $g(y)=y$ together with the mean squared error loss. For binary classification, we use the logit link $g(y)=\log\frac{y}{1-y}$ together with the binary cross-entropy loss.

\subsection{Symmetric Bigaussian Shape Function}

In this study, we employ Bigaussian (bivariate Gaussian) functions as shape functions.
We first introduce the symmetric Bigaussian shape function, and then describe the asymmetric version in the next section.

Letting $\boldsymbol{z}=[z_1,z_2]^T$ denote the input vector of the symmetric Bigaussian shape function $f_k$, we have:
\begin{eqnarray}
f_k(\boldsymbol{z}) =
a_k \exp\left(-\frac{1}{2}(\boldsymbol{z}-\boldsymbol{b_k})^T C_k^{-1}(\boldsymbol{z}-\boldsymbol{b_k})\right), \label{eq:symm_gauss}
\end{eqnarray}
where $a_k$ denotes the height of the Bigaussian, $\boldsymbol{b_k}$ represents its center position, and $C_k$ is its covariance matrix.
To clarify the shape of $f_k$ expressed by Eq.~(\ref{eq:symm_gauss}), 
we rewrite it using the eigenvalues of $C_k$. 

For the two-variable Gaussian case, let $(\lambda_{k,1}, \boldsymbol{v_{k,1}})$ and $(\lambda_{k,2}, \boldsymbol{v_{k,2}})$ denote the eigenvalue-eigenvector pairs of $C_k$.
When each eigenvector is normalized to unit length, the eigenvectors $\boldsymbol{v_{k,1}}, \boldsymbol{v_{k,2}}$ can be represented as:
\begin{eqnarray}
\begin{bmatrix}
\boldsymbol{v_{k,1}} & \boldsymbol{v_{k,2}}
\end{bmatrix}
= \begin{bmatrix}
\cos \theta_k & -\sin \theta_k \\
\sin \theta_k & \cos \theta_k
\end{bmatrix}.
\end{eqnarray}
$C_k$ can be expressed through eigen-decomposition as follows:
\begin{eqnarray}
C_k = \begin{bmatrix}
\boldsymbol{v_{k,1}} & \boldsymbol{v_{k,2}}
\end{bmatrix}
\begin{bmatrix}
\lambda_{k,1} & 0 \\
0 & \lambda_{k,2}
\end{bmatrix}
\begin{bmatrix}
\boldsymbol{v_{k,1}} & \boldsymbol{v_{k,2}}
\end{bmatrix}^T.
\end{eqnarray}
Therefore, 
the function $f_k$ can be rewritten as:
\begin{eqnarray}
f_k(\boldsymbol{z}) &=& a_k \exp\left(-\frac{1}{2}\left\|
\begin{bmatrix}
\frac{1}{\sqrt{\lambda_{k,1}}} & 0 \\
0 & \frac{1}{\sqrt{\lambda_{k,2}}}
\end{bmatrix}
\begin{bmatrix}
\cos \theta_k & \sin \theta_k \\[6pt]
-\sin \theta_k & \cos \theta_k
\end{bmatrix}
(\boldsymbol{z}-\boldsymbol{b_k})\right\|^2 \right),
\label{eq:symm_gauss3}
\end{eqnarray}
where all parameters indexed by $k$ are estimated through training. $\theta_k$ represents the rotation angle of the ellipse (contour lines of the Bigaussian function) with respect to the original coordinate axes ($z_1, z_2$). $\sqrt{\lambda_{k,1}}$ and $\sqrt{\lambda_{k,2}}$ correspond to the scaling factors along the principal axes of the ellipse.

\subsection{Asymmetric Bigaussian Shape Function}

To enable more diverse shapes while maintaining interpretability, we allow the variance of the Gaussian function to differ along the positive and negative directions of each principal axis.
To this end, we introduce a function $\phi$ that applies different scaling factors depending on whether the input is positive or negative along each axis.
The function $\phi$ is defined as follows:
\begin{eqnarray}
\phi(h; \lambda^{+}, \lambda^{-}) =
\begin{cases}
\lambda^{+} h & (h \geq 0) \\
\lambda^{-} h & (h < 0).
\end{cases}
\end{eqnarray}

\begin{figure}[t]
    \centering
    \includegraphics[scale=0.7]{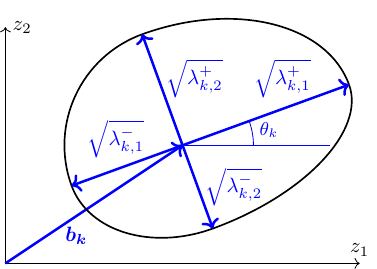}
    \caption{Relationship between learned parameters and the shape of the Bigaussian function. The vector $\boldsymbol{b_k}$ represents the center coordinates of the Bigaussian. The angle $\theta_k$ denotes the rotation of the first eigenvector relative to the $z_1$-axis. The parameters $\sqrt{\lambda_{k,1}^{+}}$ and $\sqrt{\lambda_{k,1}^{-}}$ represent the scaling factors along the first eigenvector in the positive and negative directions, respectively, relative to a standard Gaussian (i.e., one with identity covariance matrix). Similarly, $\sqrt{\lambda_{k,2}^{+}}$ and $\sqrt{\lambda_{k,2}^{-}}$ represent scaling factors along the second eigenvector in the positive and negative directions, respectively.}
    \label{fig:asymgaussian_example}
\end{figure}

Next, we modify the eigenvalues $\lambda_{k,1}, \lambda_{k,2}$ of the covariance matrix used in Eq.~(\ref{eq:symm_gauss3}). We assign different values depending on whether the input coordinate is positive or negative along each eigenvector direction.
Specifically, we use $\lambda_{k,1}^{+}$ if the input along the first eigenvector $\boldsymbol{v}_{k,1}$ is non-negative, and $\lambda_{k,1}^{-}$ if it is negative. 
We define $\lambda_{k,2}^{+}$ and $\lambda_{k,2}^{-}$ similarly.
With these adjustments, the resulting asymmetric Bigaussian shape function is given by:
\begin{eqnarray}
f_k(\boldsymbol{z}) =
a_k \exp\left(-\frac{1}{2}\left\|
\begin{bmatrix}
\phi(h_1; \frac{1}{\sqrt{\lambda_{k,1}^{+}}},\frac{1}{\sqrt{\lambda_{k,1}^{-}}}) \\
\phi(h_2; \frac{1}{\sqrt{\lambda_{k,2}^{+}}},\frac{1}{\sqrt{\lambda_{k,2}^{-}}})
\end{bmatrix}
\right\|^2 \right), \label{eq:asymm_gauss}
\end{eqnarray}
where
\begin{eqnarray}
\begin{bmatrix}
h_1 \\
h_2
\end{bmatrix}
=
\begin{bmatrix}
\cos \theta_k & \sin \theta_k \\
-\sin \theta_k & \cos \theta_k
\end{bmatrix}(\boldsymbol{z}-\boldsymbol{b_k}).
\end{eqnarray}
Figure~\ref{fig:asymgaussian_example} illustrates the geometric interpretation of each parameter.
The four $\lambda$ parameters represent scaling factors of the ellipse in the positive and negative directions along each eigenvector, relative to a standard Gaussian function (i.e., a Gaussian with an identity covariance matrix).

\subsection{Model Training}
\label{sec:model_training}

We implement the BAM given by Eq.~(\ref{eq:bam_score}) as a neural network, and estimate its parameters via backpropagation. The shape functions within the model are defined by Eq.~(\ref{eq:asymm_gauss}).
To enforce the constraint that eigenvalues of the covariance matrices must be positive, we introduce an additional parameter $\lambda^{\prime}$ and define each eigenvalue $\lambda$ as follows:
\begin{eqnarray}
\lambda = \exp(s \lambda^{\prime}),
\end{eqnarray}
where $s$ controls the magnitude of changes in eigenvalues, and we set $s=1.5$ throughout this paper. In our experiments, the initial value of $\lambda^{\prime}$ is set to $-1.0$.
The amplitude $a_k$ is free to take negative values, allowing each component to represent either a ``peak'' or a ``valley'' in the contour map; we initialize $a_k$ and $\boldsymbol{b}_k$ from a standard normal distribution, and set the rotation angle $\theta_k$ to zero.

While the input features to each shape function are specified by the tuple $\mathcal{I}$ in Eq.~(\ref{eq:feature_indecies}), we introduce a differentiable selection mechanism to enable the neural network to automatically learn these feature selections.
We express each element of the input vector $\boldsymbol{z}=[z_1,z_2]^T$ for a bivariate shape function as a weighted average of all available features.
Through training, the weighting coefficients are optimized to converge to a one-hot vector, thereby selecting a single specific feature.
Following NODE-GAM \cite{chang22}, we adopt a temperature annealing parameter $T_e$, which gradually enforces these weight coefficients to become a one-hot vector as training progresses.
The input vector $\boldsymbol{z}$ is computed as:
\begin{eqnarray}
\boldsymbol{z} = \mathrm{entmax}_{\alpha}(W/T_e)\boldsymbol{x}, \label{eq:entmax}
\end{eqnarray}
where $W \in \mathbb{R}^{2 \times m}$ is a weight matrix specific to each component, and the $\mathrm{entmax}$ activation function \cite{peters2019} with $\alpha=1.5$ is applied separately to each row of $W/T_e$.
The entmax function induces sparsity in each weight vector. 
The temperature parameter $T_e$ is annealed gradually from an initial value $T_{\max}=1.0$ to a final value $T_{\min}=0.001$.
Given the maximum number of epochs $E$, the temperature at epoch $e$, denoted by $T_e$, is computed as:
\begin{eqnarray}
T_e = \exp\left( \log T_{\max} - \frac{e}{E}\left(\log T_{\max} - \log T_{\min}\right)\right),
\end{eqnarray}
where epochs are indexed from $e=0$ to $e=E$.
After this annealing stage, we fix the selection of each row of $W$ to the feature with the largest weight (i.e., replace the softened selection with its argmax) and continue training the remaining parameters for a second stage, so that the Gaussian parameters are refined under the fixed feature selection. Because entmax is applied to each row of $W/T_e$ independently, the two rows may converge to the same feature, in which case the component reduces to a univariate shape function.

\subsection{Multiple Restart Strategy}

The optimization process of our proposed method is sensitive to initialization.
To mitigate this sensitivity, we adopted a multiple restart strategy, similar to the standard practice in K-Means clustering \cite{scikit-learn}.
Specifically, we executed the training process ten times with different random initializations for each fold and selected the model with the lowest training loss.
Selecting the best model based on training loss (rather than validation loss) is unlikely to cause overfitting, due to the model's simplicity.

\section{Evaluation Results}
\label{sec:evaluation}

This section evaluates the performance of the proposed BAM.
First, we compare BAM with other predictive models. 
Next, we compare the visualization of BAM with those of existing methods.
We then investigate how the number of Gaussian components influences model performance. 

\subsection{Datasets}

We evaluated our method using three regression and three binary classification datasets. For regression, we used: California Housing \cite{ca_housing1997} (CA Housing, 20,640 samples, 8 numerical features); Bike Sharing \cite{bike_sharing2013} (Bikeshare, 17,379 samples, 13 numerical features); and Wine Quality \cite{wine_quality2009} (Wine, white wine subset, 4,898 samples, 11 numerical features). For binary classification, we used: Titanic \cite{titanic2012} (712 samples after dropping missing values, `sex' and `embarked' as categorical, others numerical); Adult \cite{adult1996} (13,379 samples, 8 categorical features); and Diabetes Health Indicators \cite{diabetes2015} (Diabetes, balanced version, 70,692 samples, 21 numerical features).
As preprocessing, categorical variables were one-hot encoded, and numerical variables (including targets for regression) were standardized.

\subsection{Baseline Evaluation}
We evaluated predictive performance (RMSE for regression, AUC for classification) using 5-fold cross-validation.
For BAM, we evaluated a model with 15 asymmetric components, trained with the Adam optimizer (learning rate $0.001$, batch size $128$) for $150$ epochs in the annealing stage, followed by $150$ additional epochs with the feature selection fixed (the second stage described in Sect.~\ref{sec:model_training}).

The comparison models used in this evaluation were as follows.
Linear Regression/Logistic Regression (Linear), Decision Tree (CART), and XGBoost were used with their default settings.
For Explainable Boosting Machine (EBM), we used InterpretML. We set \texttt{interactions=0} for the GAM configuration, while using the default setting (\texttt{interactions="5x"}) for the GA\textsuperscript{2}M configuration.
For NODE-GAM, we used the authors' implementation for both GAM and GA\textsuperscript{2}M, removing only the default preprocessing.
For GRAND-SLAMIN', we used the official implementation with its default architecture. As with the other baselines, we did not tune hyperparameters: we replaced the Optuna-based hyperparameter search with a standard mini-batch training loop (Adam, learning rate $5 \times 10^{-3}$, batch size 128, 300 epochs). To stabilize training on high-dimensional inputs such as the one-hot encoded Adult dataset, we initialized the selection gates so that all gates start near zero and scaled the regularization coefficients inversely with the number of candidate terms. 
Finally, as a Deep Neural Network (DNN) baseline, we trained a 5-layer fully connected network (64-64-64-32-1 units) with 0.2 dropout using Adam (learning rate 0.001, batch size 128) for 150 epochs.

Table~\ref{table:performance_comp} presents the predictive performance.
BAM achieves accuracy comparable to standard GAMs and approaches that of GA\textsuperscript{2}Ms on some datasets.
On Bikeshare, BAM even outperforms the GAMs by a large margin and comes close to the GA\textsuperscript{2}Ms.
This is likely because Bikeshare contains strong interactions between features, which univariate shape functions cannot capture.

\begin{table}[t]
\caption{Comparison of predictive performance on multiple datasets (Mean $\pm$ Std). BAM denotes the proposed model with $n_c = 15$ asymmetric Gaussian components.}
\label{table:performance_comp}
\centering
\scriptsize
\renewcommand{\arraystretch}{0.92}
\setlength{\tabcolsep}{3pt}
\begin{tabular}{lcccccc}
\toprule
\textbf{Model}
& \thead{CA Housing \\ (RMSE$\downarrow$)}
& \thead{Bikeshare \\ (RMSE$\downarrow$)}
& \thead{Wine \\ (RMSE$\downarrow$)}
& \thead{Titanic \\ (AUC$\uparrow$)}
& \thead{Adult \\ (AUC$\uparrow$)}
& \thead{Diabetes \\ (AUC$\uparrow$)} \\
\midrule
Linear & \pmsd{0.728}{0.017} & \pmsd{141.907}{2.022} & \pmsd{0.755}{0.030} & \pmsd{0.855}{0.039} & \pmsd{0.907}{0.003} & \pmsd{0.825}{0.003} \\
CART & \pmsd{0.720}{0.010} & \pmsd{58.550}{1.849} & \pmsd{0.821}{0.026} & \pmsd{0.731}{0.024} & \pmsd{0.749}{0.001} & \pmsd{0.656}{0.006} \\
\midrule
EBM (GAM) & \pmsd{0.550}{0.013} & \pmsd{99.835}{1.465} & \pmsd{0.698}{0.020} & \pmsd{0.857}{0.041} & \pmsd{0.928}{0.003} & \pmsd{0.829}{0.002} \\
NODE-GAM (GAM) & \pmsd{0.597}{0.015} & \pmsd{100.234}{1.486} & \pmsd{0.704}{0.021} & \pmsd{0.846}{0.045} & \pmsd{0.918}{0.002} & \pmsd{0.829}{0.002} \\
\midrule
BAM (ours) & \pmsd{0.589}{0.020} & \pmsd{65.990}{1.849} & \pmsd{0.721}{0.021} & \pmsd{0.869}{0.040} & \pmsd{0.912}{0.002} & \pmsd{0.824}{0.004} \\
\midrule
EBM (GA\textsuperscript{2}M) & \pmsd{0.462}{0.012} & \pmsd{49.428}{0.852} & \pmsd{0.667}{0.014} & \pmsd{0.865}{0.036} & \pmsd{0.929}{0.003} & \pmsd{0.831}{0.002} \\
NODE-GAM (GA\textsuperscript{2}M) & \pmsd{0.538}{0.024} & \pmsd{57.637}{3.380} & \pmsd{0.685}{0.018} & \pmsd{0.844}{0.034} & \pmsd{0.919}{0.002} & \pmsd{0.832}{0.002} \\
GRAND-SLAMIN' & \pmsd{0.564}{0.019} & \pmsd{57.975}{2.105} & \pmsd{0.704}{0.018} & \pmsd{0.868}{0.030} & \pmsd{0.914}{0.003} & \pmsd{0.829}{0.002} \\
\midrule
DNN & \pmsd{0.512}{0.012} & \pmsd{56.088}{1.011} & \pmsd{0.670}{0.013} & \pmsd{0.852}{0.010} & \pmsd{0.897}{0.002} & \pmsd{0.824}{0.001} \\
XGBoost & \pmsd{0.469}{0.012} & \pmsd{40.633}{1.298} & \pmsd{0.633}{0.016} & \pmsd{0.838}{0.035} & \pmsd{0.928}{0.002} & \pmsd{0.823}{0.002} \\
\bottomrule
\end{tabular}
\end{table}

\subsection{Model Visualization}
Figure~\ref{fig:visualization} visualizes the BAM model trained on the Titanic dataset, which consists of three bivariate components and a bias term. The visualized values represent logits, and their sum (plus the bias term) determines the final prediction; a positive total score indicates survival.
The decomposition reveals how the model captures key survival factors. 
We discuss each component in turn and compare it with what is historically known about Titanic survival outcomes, illustrating how BAM's visualization supports model auditing.
For the first component (class and gender), 
the Gaussian peak lies at (\texttt{sex\_male}=0, \texttt{pclass}=1), so female passengers in first class receive a strongly positive contribution. This is consistent with the view that women in first class were more likely to have been prioritized for evacuation.
The second component (age and class) identifies a strong negative contribution for adult passengers in the lower class, with the Gaussian centered in the region (\texttt{age} large, \texttt{pclass}=3). This is consistent with reported low survival rates for this group. Such a conditional structure cannot be captured by a univariate function on \texttt{age} or \texttt{pclass} alone.
The third component (single-female adjustment) captures a more subtle adjustment: the Gaussian peak lies at (\texttt{sex\_male}=0, \texttt{sibsp}=0), so single female passengers receive an additional positive contribution beyond what Component 1 already provides for women. 

\begin{figure}[t]
    \centering
    \begin{minipage}[b]{0.19\linewidth}\centering
        \includegraphics[width=\linewidth]{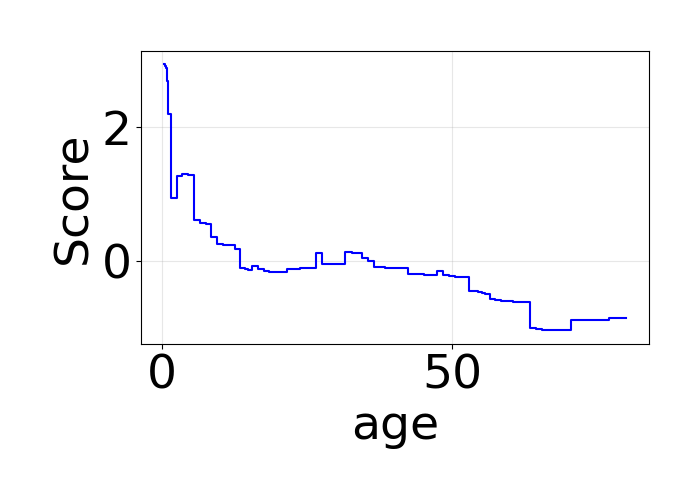}
    \end{minipage}\hfill
    \begin{minipage}[b]{0.19\linewidth}\centering
        \includegraphics[width=\linewidth]{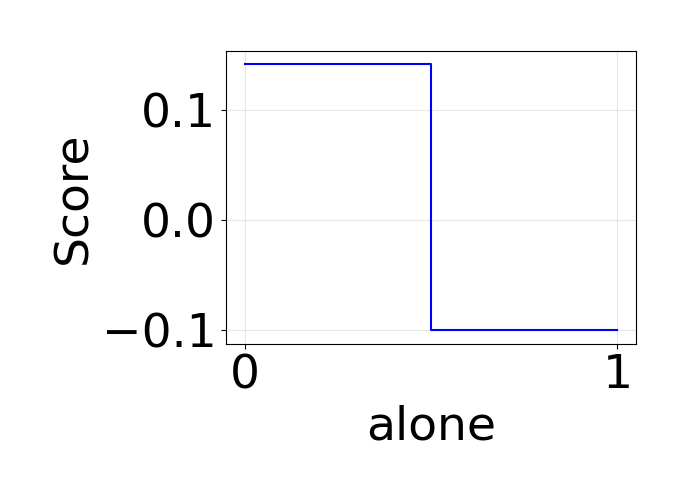}
    \end{minipage}\hfill
    \begin{minipage}[b]{0.19\linewidth}\centering
        \includegraphics[width=\linewidth]{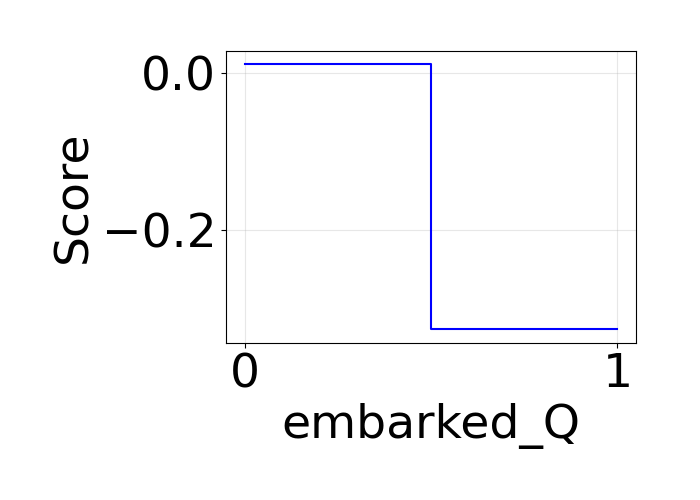}
    \end{minipage}\hfill
    \begin{minipage}[b]{0.19\linewidth}\centering
        \includegraphics[width=\linewidth]{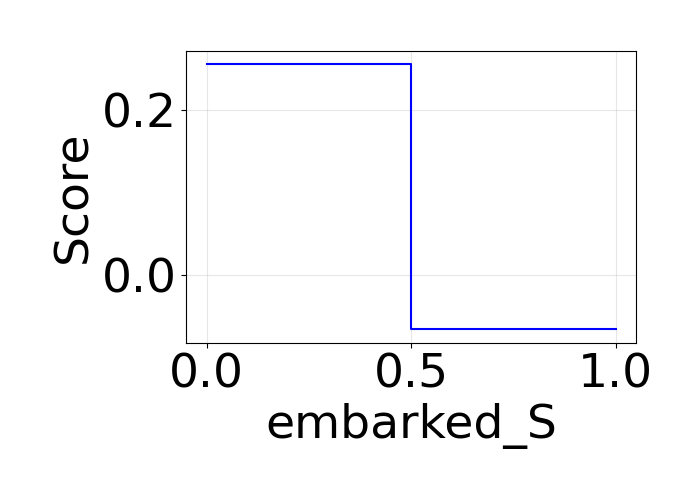}
    \end{minipage}\hfill
    \begin{minipage}[b]{0.19\linewidth}\centering
        \includegraphics[width=\linewidth]{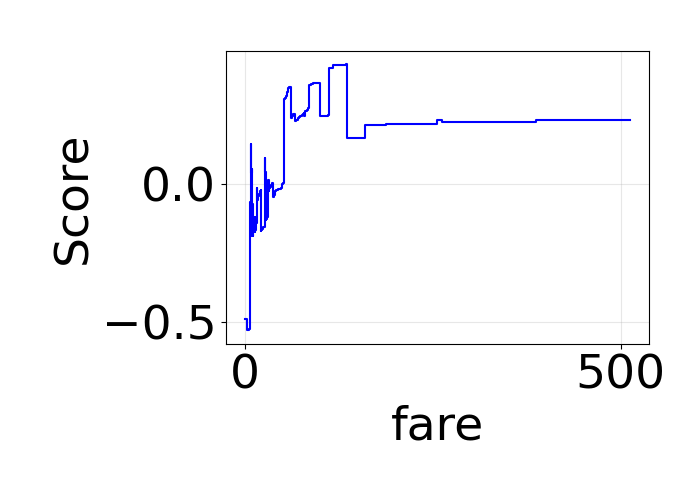}
    \end{minipage}

    \begin{minipage}[b]{0.19\linewidth}\centering
        \includegraphics[width=\linewidth]{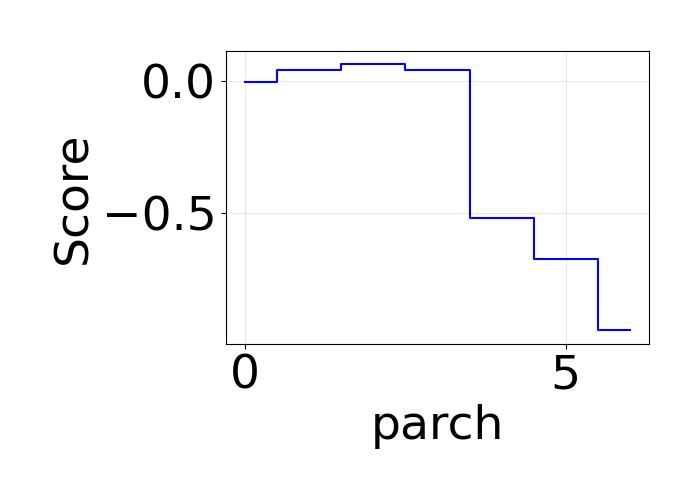}
    \end{minipage}\hfill
    \begin{minipage}[b]{0.19\linewidth}\centering
        \includegraphics[width=\linewidth]{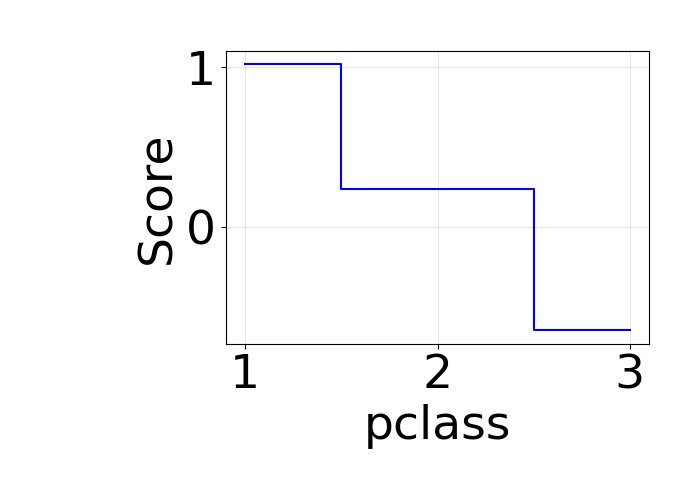}
    \end{minipage}\hfill
    \begin{minipage}[b]{0.19\linewidth}\centering
        \includegraphics[width=\linewidth]{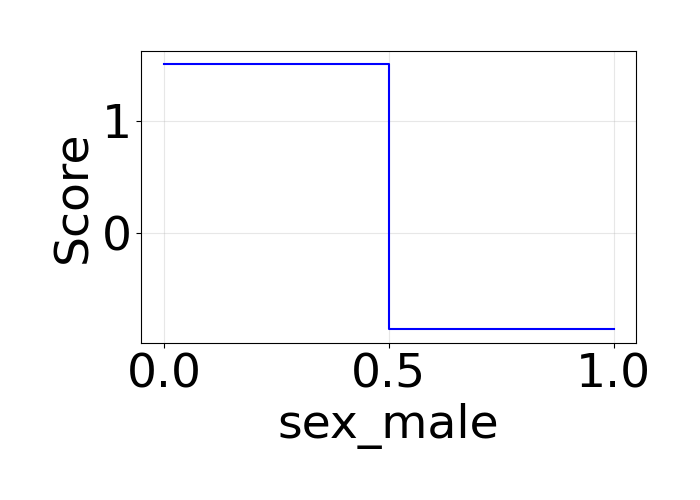}
    \end{minipage}\hfill
    \begin{minipage}[b]{0.19\linewidth}\centering
        \includegraphics[width=\linewidth]{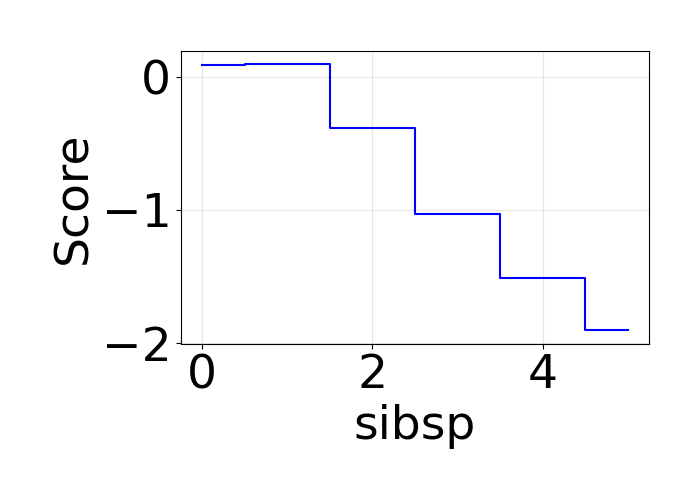}
    \end{minipage}\hfill
    \begin{minipage}[b]{0.19\linewidth}\rule{0mm}{1mm}\end{minipage}

    \begin{minipage}[b]{0.19\linewidth}\centering
        \includegraphics[width=\linewidth]{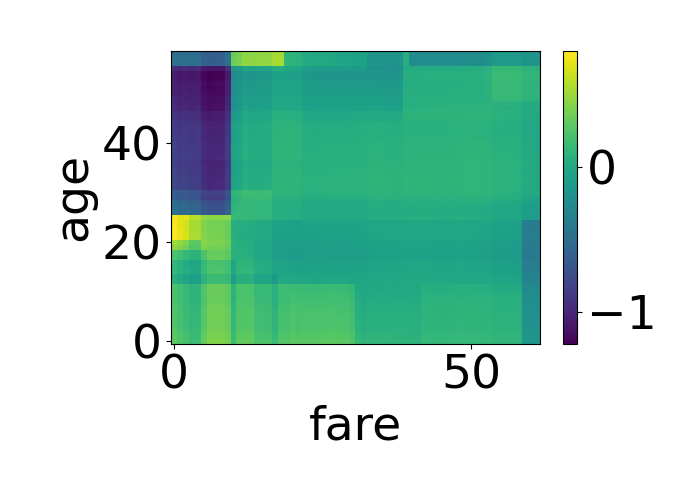}
    \end{minipage}\hfill
    \begin{minipage}[b]{0.19\linewidth}\centering
        \includegraphics[width=\linewidth]{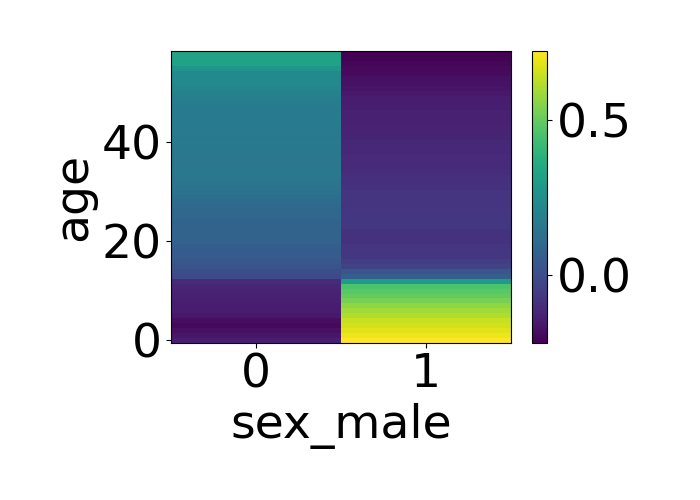}
    \end{minipage}\hfill
    \begin{minipage}[b]{0.19\linewidth}\centering
        \includegraphics[width=\linewidth]{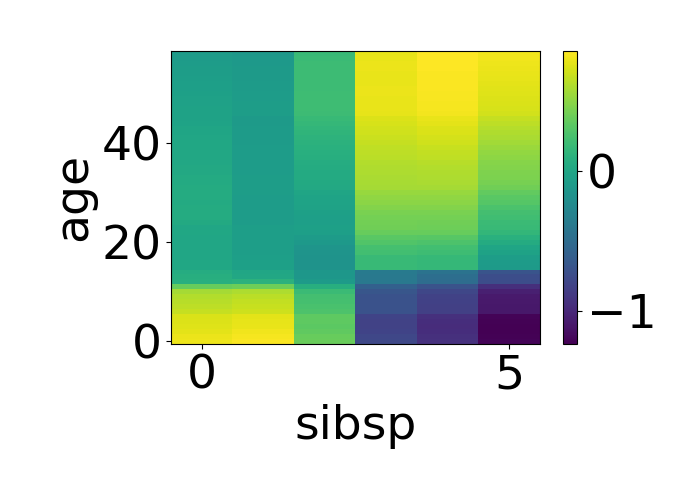}
    \end{minipage}\hfill
    \begin{minipage}[b]{0.19\linewidth}\centering
        \includegraphics[width=\linewidth]{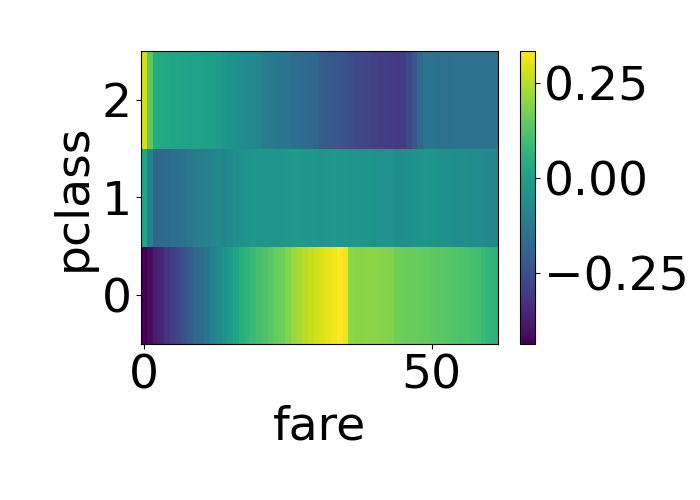}
    \end{minipage}\hfill
    \begin{minipage}[b]{0.19\linewidth}\centering
        \includegraphics[width=\linewidth]{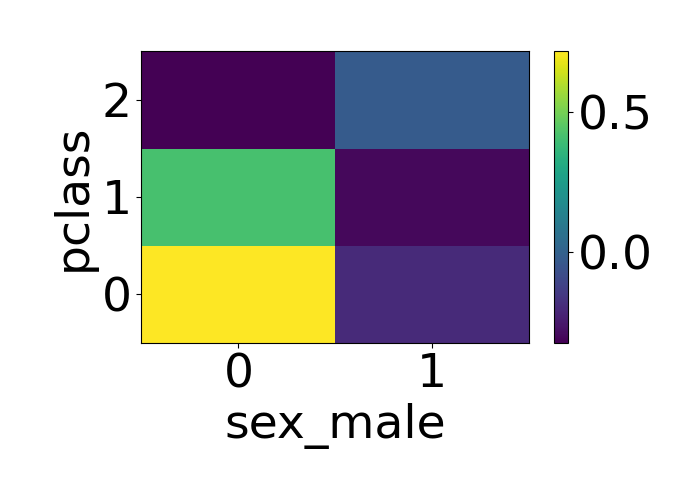}
    \end{minipage}

    \caption{Visualization of EBM (GA\textsuperscript{2}M) shape functions on the Titanic dataset. The first two rows show univariate shape functions; the last row shows pairwise interaction shape functions. The number of pairwise interactions was reduced from the InterpretML default (\texttt{interactions="5x"}) to \texttt{"0.5x"} (half the number of features) for this figure only.}
    \label{fig:ebm_visualization}
\end{figure}

\begin{figure}[t]
    \centering
    \begin{minipage}[b]{0.19\linewidth}\centering
        \includegraphics[width=\linewidth]{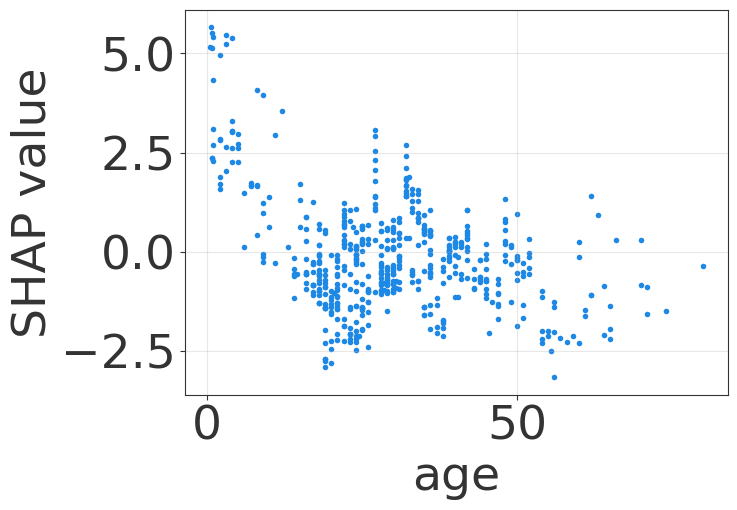}
    \end{minipage}\hfill
    \begin{minipage}[b]{0.19\linewidth}\centering
        \includegraphics[width=\linewidth]{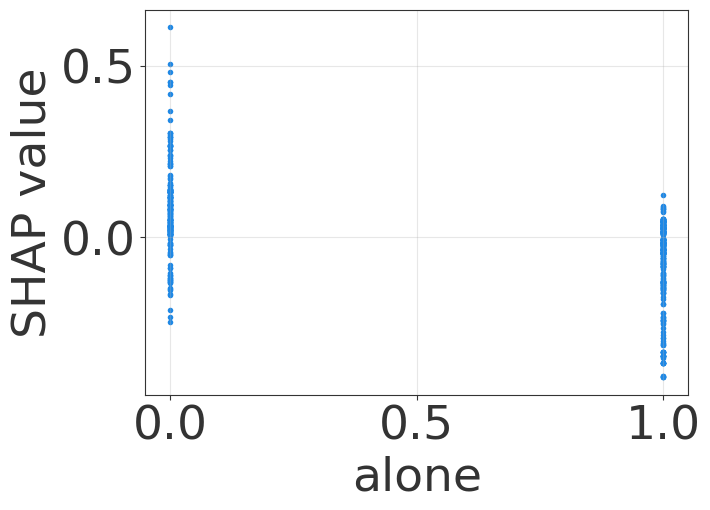}
    \end{minipage}\hfill
    \begin{minipage}[b]{0.19\linewidth}\centering
        \includegraphics[width=\linewidth]{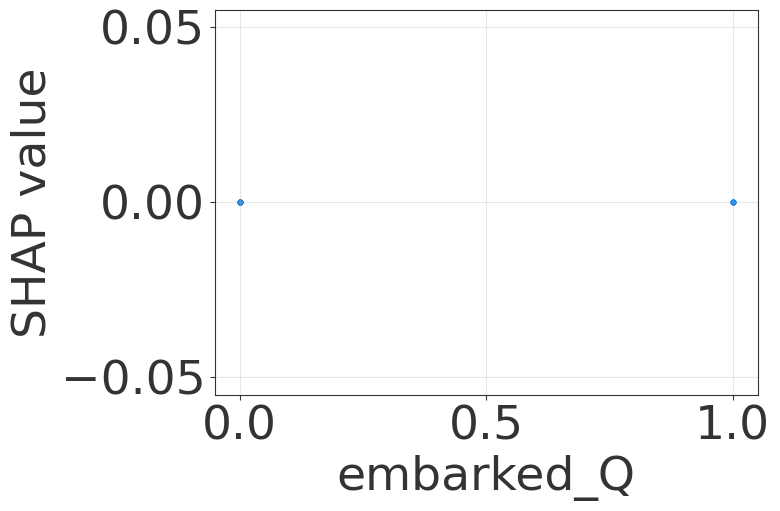}
    \end{minipage}\hfill
    \begin{minipage}[b]{0.19\linewidth}\centering
        \includegraphics[width=\linewidth]{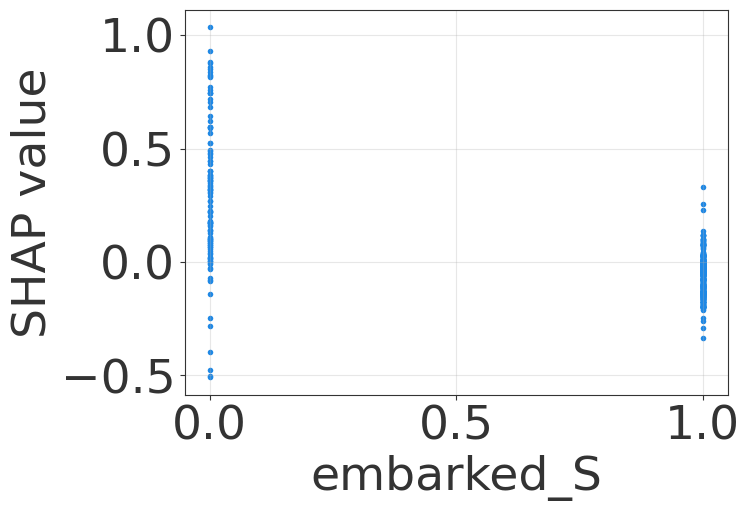}
    \end{minipage}\hfill
    \begin{minipage}[b]{0.19\linewidth}\centering
        \includegraphics[width=\linewidth]{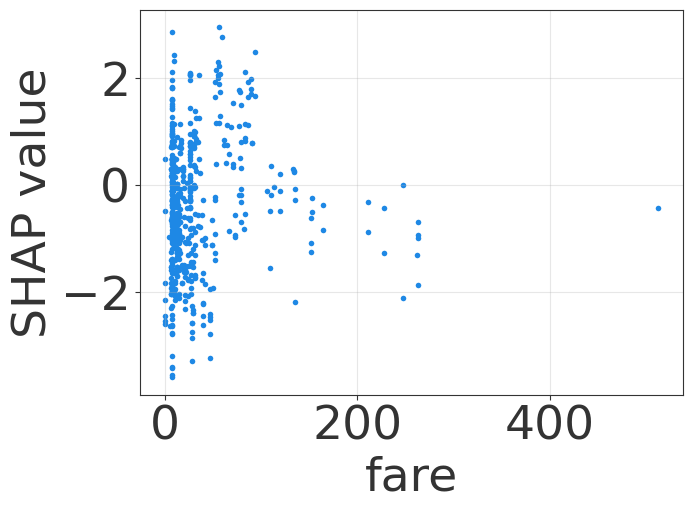}
    \end{minipage}

    \begin{minipage}[b]{0.19\linewidth}\centering
        \includegraphics[width=\linewidth]{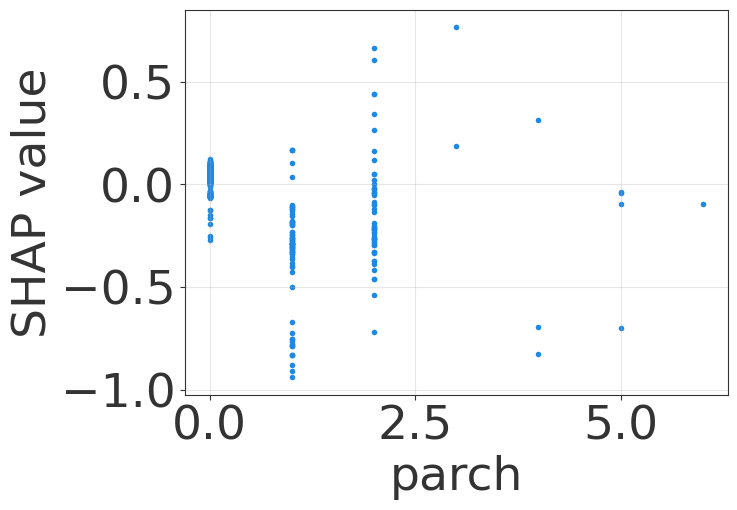}
    \end{minipage}\hfill
    \begin{minipage}[b]{0.19\linewidth}\centering
        \includegraphics[width=\linewidth]{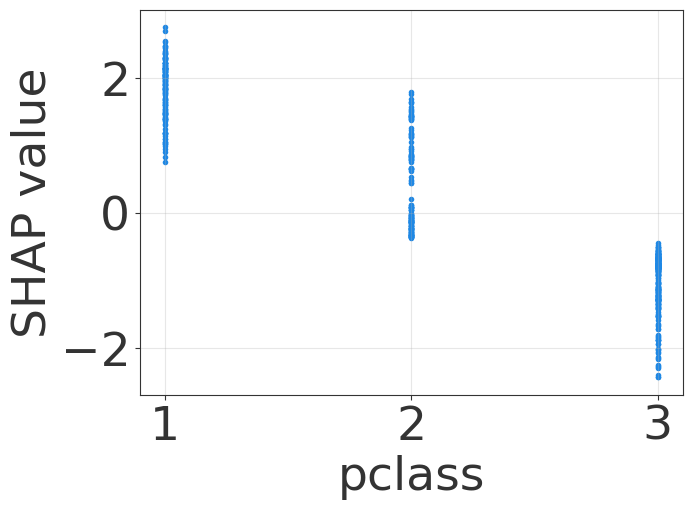}
    \end{minipage}\hfill
    \begin{minipage}[b]{0.19\linewidth}\centering
        \includegraphics[width=\linewidth]{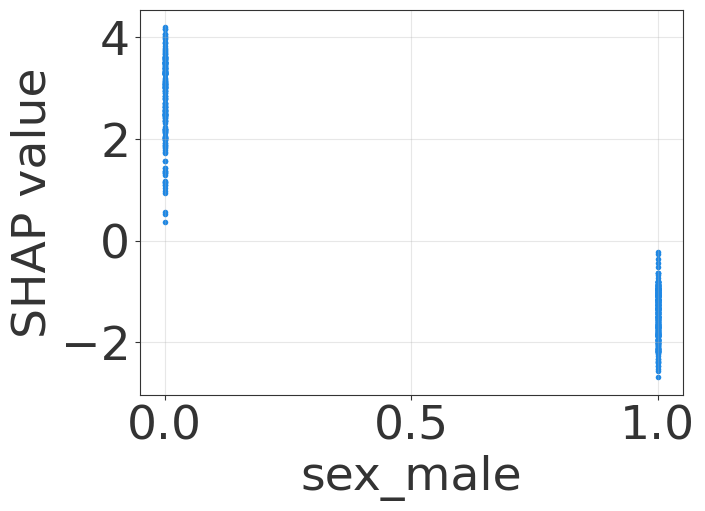}
    \end{minipage}\hfill
    \begin{minipage}[b]{0.19\linewidth}\centering
        \includegraphics[width=\linewidth]{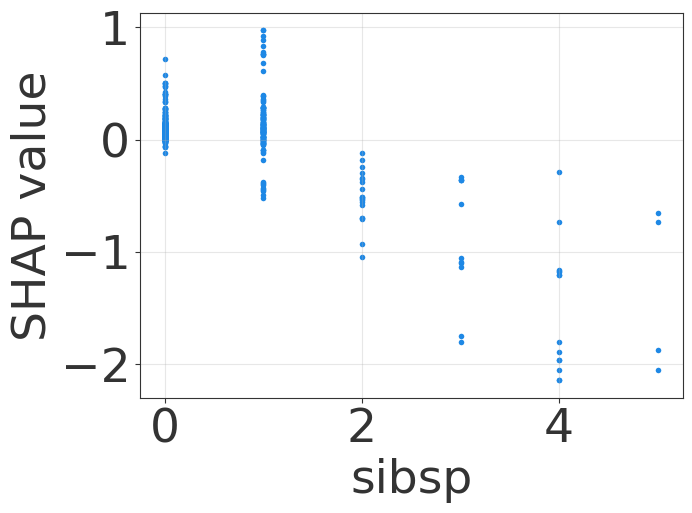}
    \end{minipage}\hfill
    \begin{minipage}[b]{0.19\linewidth}\rule{0mm}{1mm}\end{minipage}
    \caption{SHAP dependence plots for the XGBoost model on the Titanic dataset. Each plot shows the relationship between a feature value and its SHAP value.}
    \label{fig:shap_visualization}
\end{figure}

For comparison, we visualize the results of EBM (GA\textsuperscript{2}M) and SHAP in Figs.~\ref{fig:ebm_visualization} and \ref{fig:shap_visualization}, respectively.
Figure~\ref{fig:ebm_visualization} shows the visualization of EBM. Note that for this visualization, we limited the number of interactions to half the number of features to avoid an excessive number of plots (\texttt{interactions="0.5x"}). Even with this reduction, the large number of plots makes it difficult to grasp the overall picture. Furthermore, the learned functions for `age' and `fare' exhibit irregular shapes, making it difficult to grasp underlying trends.

Figure~\ref{fig:shap_visualization} shows the SHAP dependence plots of the XGBoost model. In each plot, the horizontal axis represents the value of a feature, and the vertical axis represents its SHAP value, i.e., the contribution of that feature to the model's output. A notable observation is that, even at the same feature value, the SHAP value varies considerably across samples, producing a vertical spread of points rather than a single curve. This spread indicates that the contribution of a feature is not determined by its own value alone, but also depends on the values of other features.
However, the dependence plot does not reveal which other features are responsible for this variation, nor how they jointly shape the contribution. As a result, 
the underlying logic that links feature values to the model's output remains opaque.

\begin{figure}[t]
    \centering
    \includegraphics[width=\linewidth]{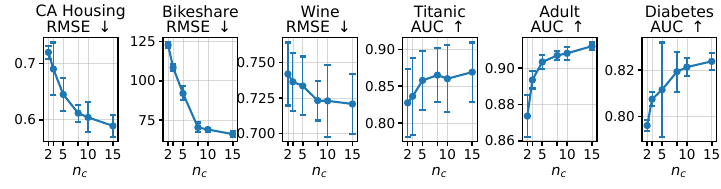}
    \caption{Relationship between the number of components and predictive performance (mean $\pm$ std over 5 folds).}
    \label{fig:num_components}
\end{figure}

\subsection{Effect of the Number of Gaussian Components}

Figure~\ref{fig:num_components} illustrates how predictive performance changes as the number of components varies. All other experimental settings remain consistent with the baseline evaluation.
The results show that increasing the number of components generally improves predictive performance.
This finding suggests that the number of components serves as an effective tunable parameter, enabling practitioners to balance accuracy against the effort of inspecting the components according to their specific needs.

\section{Conclusion}
\label{sec:conclusion}

We addressed the gap between the nominal interpretability of GA\textsuperscript{2}Ms and what their shape functions actually allow a practitioner to inspect. We proposed Bivariate Additive Models (BAMs), a class of GA\textsuperscript{2}Ms in which every pairwise shape function is constrained to a simple parametric form, instantiated as an asymmetric bivariate Gaussian. BAMs select the feature pair of each component via a differentiable selection mechanism, and each component is visualizable as a unimodal contour map.
Experiments on six tabular benchmarks showed that BAM attains predictive accuracy comparable to standard GAMs and approaches that of GA\textsuperscript{2}Ms. 
We also showed the trade-off between accuracy and the number of components to inspect.
BAM is positioned not as the highest-accuracy model but as an auditable additive model: it is well suited to tabular prediction tasks where each feature pair's contribution must be transparently inspectable, such as industrial sensing applications in which engineers need to identify and share the dominant variable pairs driving outcomes.
Future work includes exploring alternative simple forms for the shape function (e.g., monotonic functions), investigating identifiability and concurvity~\cite{siems23} among components, and clarifying the relationship between dataset characteristics and optimal model parameters such as the number of components.

\begin{credits}
\subsubsection{\ackname} This work was supported by JSPS KAKENHI Grant Number JP24K15081.
%
\end{credits}
%
\bibliographystyle{splncs04}
\bibliography{xai}
\end{document}